\documentclass[10pt,twocolumn,letterpaper]{article}

\usepackage{cvpr}
\usepackage{times}
\usepackage{epsfig}
\usepackage{graphicx}
\usepackage{amsmath}
\usepackage{amssymb}
\newcommand{\old}[1]{}
\usepackage[pagebackref=true,breaklinks=true,letterpaper=true,colorlinks,bookmarks=false]{hyperref}

\cvprfinalcopy % *** Uncomment this line for the final submission

\def\cvprPaperID{892} % *** Enter the CVPR Paper ID here

\ifcvprfinal\fi
\begin{document}

%%%%%%%%% TITLE
\title{Deep Homography Estimation for Dynamic Scenes}

\author{Hoang Le \hspace{1.75cm} Feng Liu$^*$\\
\hspace{0cm}Portland State University\\
\hspace{0cm}{\tt\small \{hoanl,fliu\}@cs.pdx.edu}
% \and
% Feng Liu\\
\and
Shu Zhang\\
Google\\
{\tt\small shzhang@google.com}
\and
Aseem Agarwala\thanks{Work mostly done while Feng and Aseem were at Google Research.}\\
Adobe Research\\
{\tt\small aseem@agarwala.org}
}
\maketitle
%\thispagestyle{empty}

%%%%%%%%% ABSTRACT
\begin{abstract}

Homography estimation is an important step in many computer vision problems. Recently, deep neural network methods have shown to be favorable for this problem when compared to traditional methods. However, these new methods do not consider dynamic content in input images. They train neural networks with only image pairs that can be perfectly aligned using homographies. This paper investigates and discusses how to design and train a deep neural network that handles dynamic scenes. We first collect a large video dataset with dynamic content\footnote{Dataset: \href{https://github.com/lcmhoang/hmg-dynamics}{https://github.com/lcmhoang/hmg-dynamics}}. We then develop a multi-scale neural network and show that when properly trained using our new dataset, this neural network can already handle dynamic scenes to some extent. To estimate a homography of a dynamic scene in a more principled way, we need to identify the dynamic content. Since dynamic content detection and homography estimation are two tightly coupled tasks, we follow the multi-task learning principles and augment our multi-scale network such that it jointly estimates the dynamics masks and homographies. Our experiments show that our method can robustly estimate homography for challenging scenarios with dynamic scenes, blur artifacts, or lack of textures.
\end{abstract}

%%%%%%%%% BODY TEXT
\section{Introduction}
A homography models the global geometric transformation 
between two images. It not only directly serves as a motion model for many applications like video stabilization~\cite{gleicher2007re,matsushita2006full} and
image stitching~\cite{szeliski2007image,zhang2014parallax}, but also is used to estimate
3D motion and scene structure in algorithms, such as SLAM~\cite{engel2014lsd} and visual odometry~\cite{forster2014svo}.

\begin{figure}[t]
\begin{center}
\begin{tabular}{ccc}        
    \hspace{-0.3cm} \includegraphics[width=0.33\linewidth]{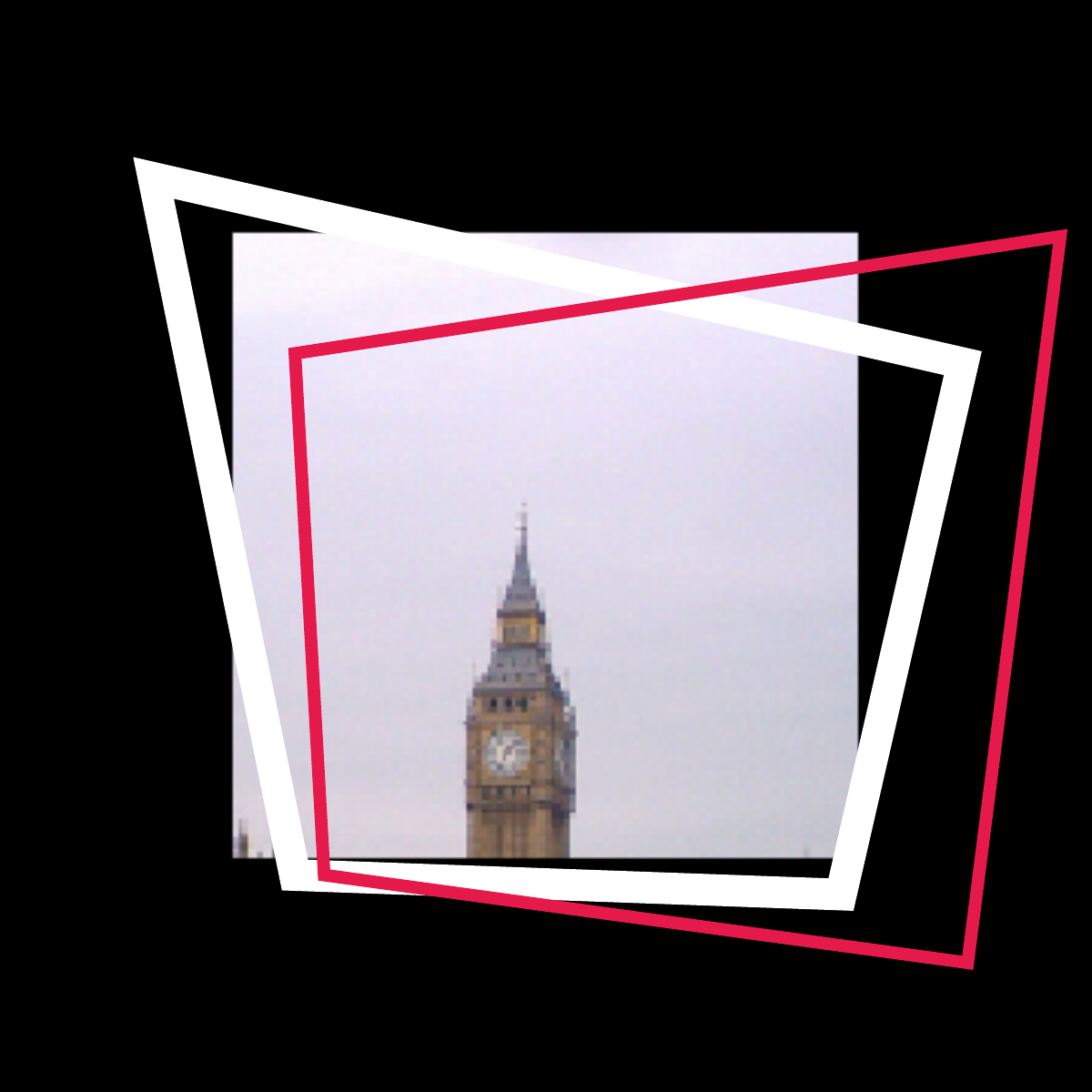}
    &
    \hspace{-0.45cm} \includegraphics[width=0.33\linewidth]{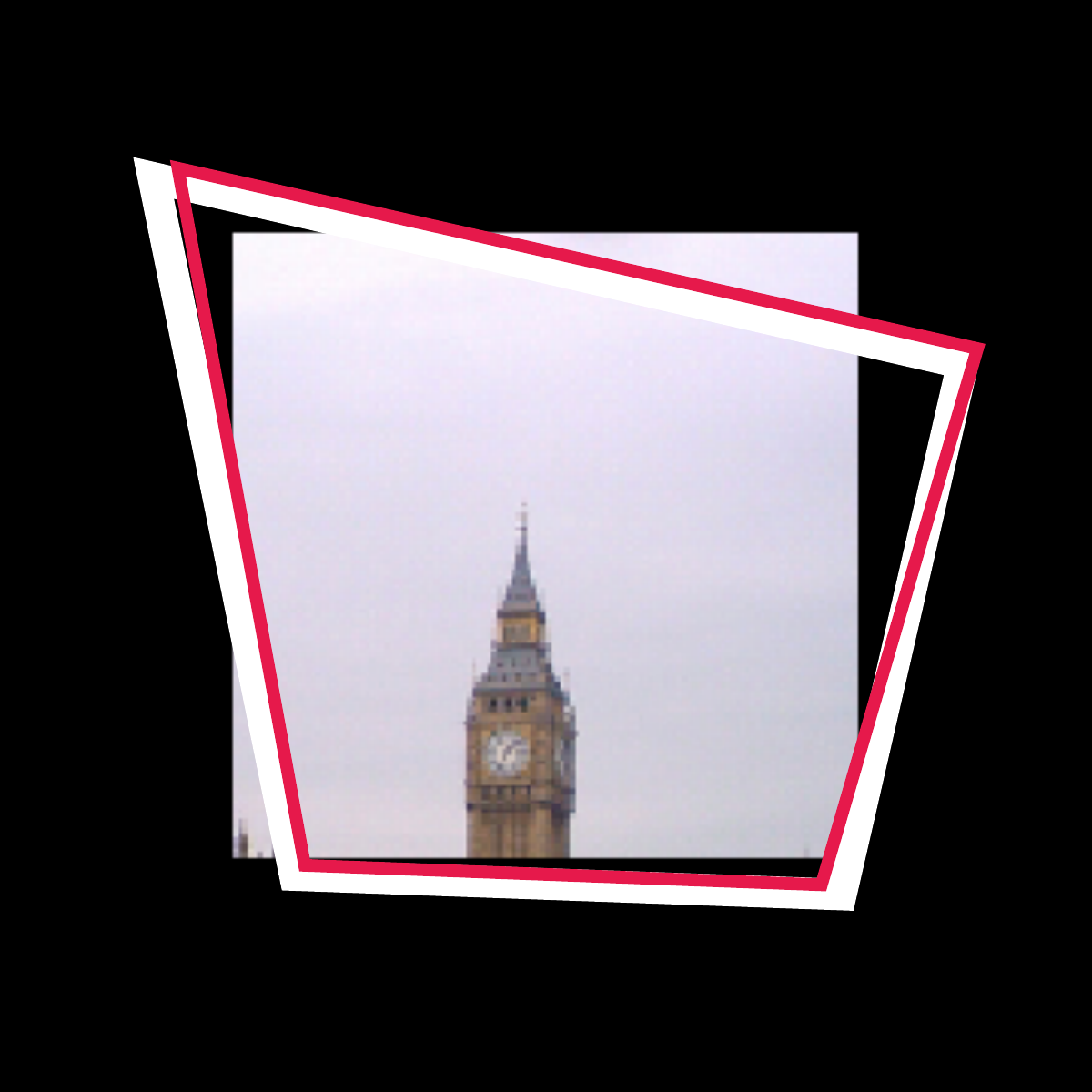}
    &
    \hspace{-0.45cm} \includegraphics[width=0.33\linewidth]{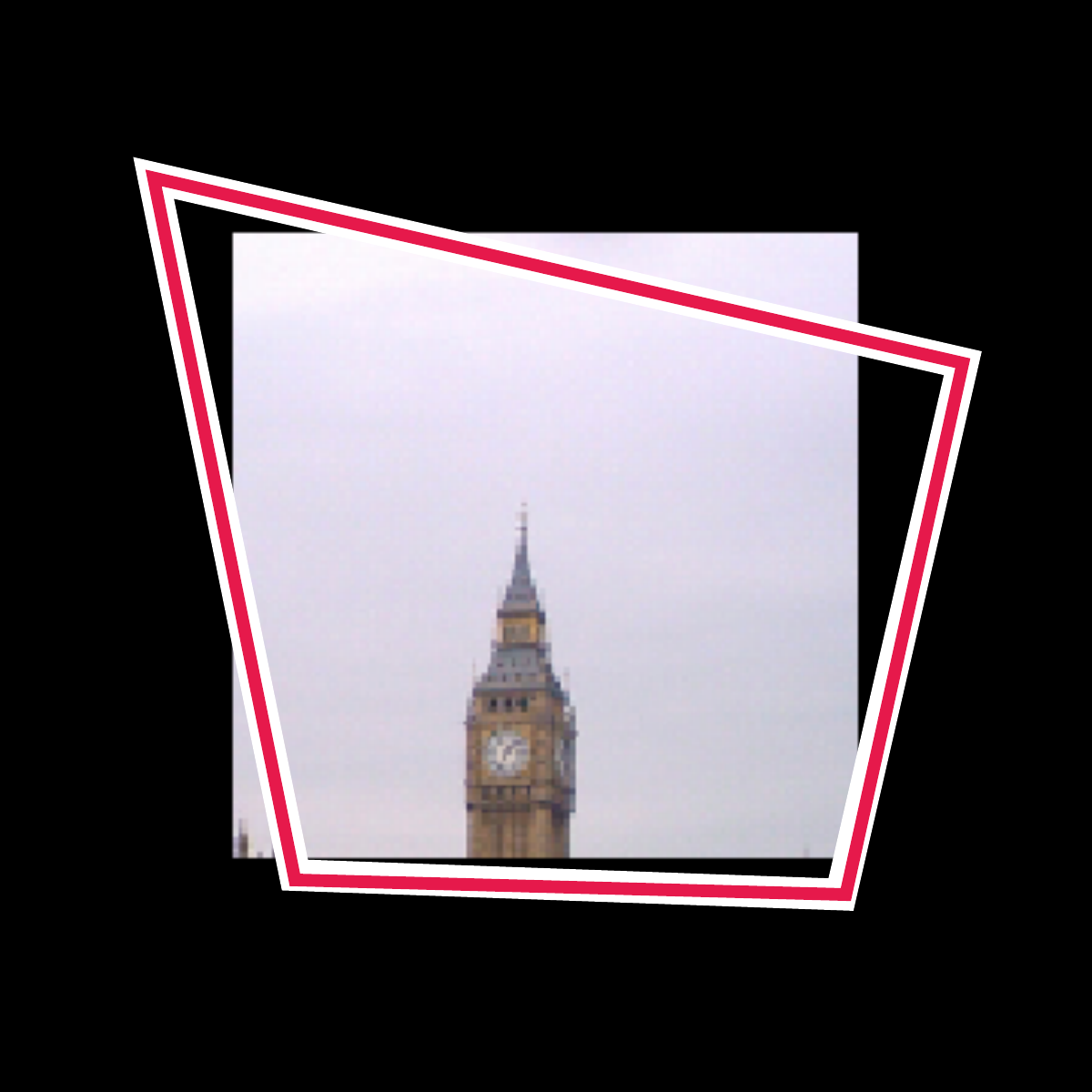}        
    \vspace{-0.5mm}
    \\
    \hspace{-0.3cm} \includegraphics[width=0.33\linewidth]{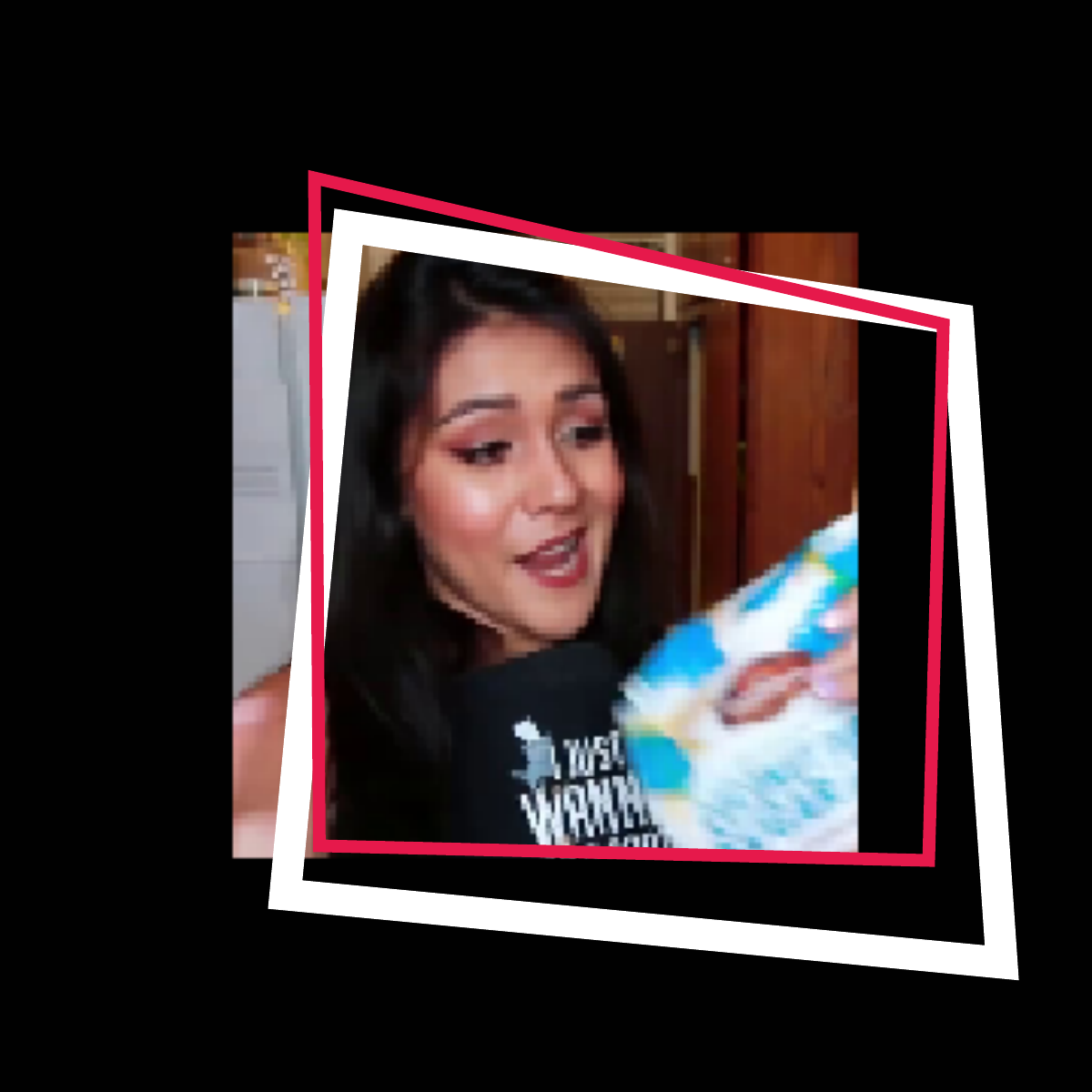}
    &
    \hspace{-0.45cm} \includegraphics[width=0.33\linewidth]{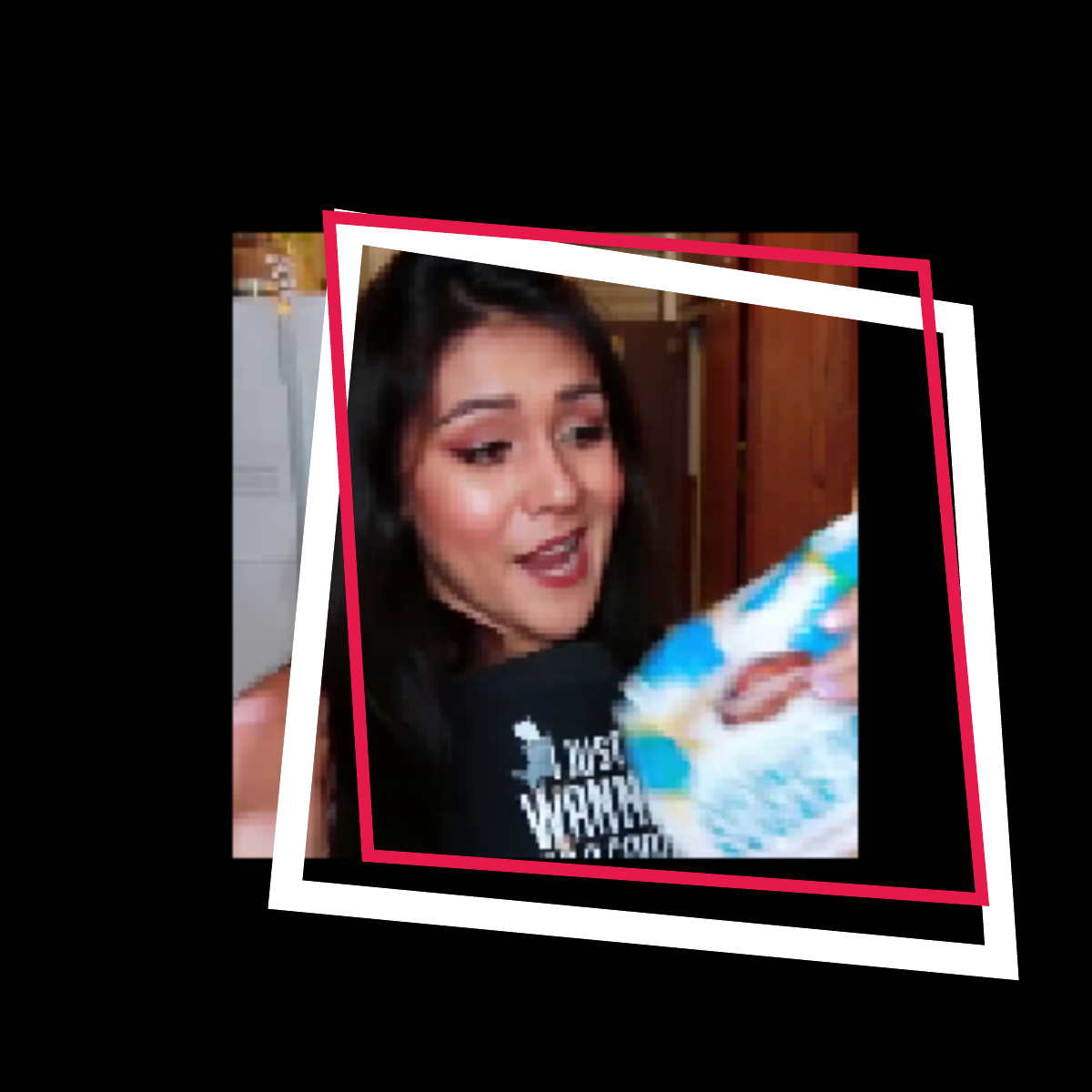}
    &
    \hspace{-0.45cm} \includegraphics[width=0.33\linewidth]{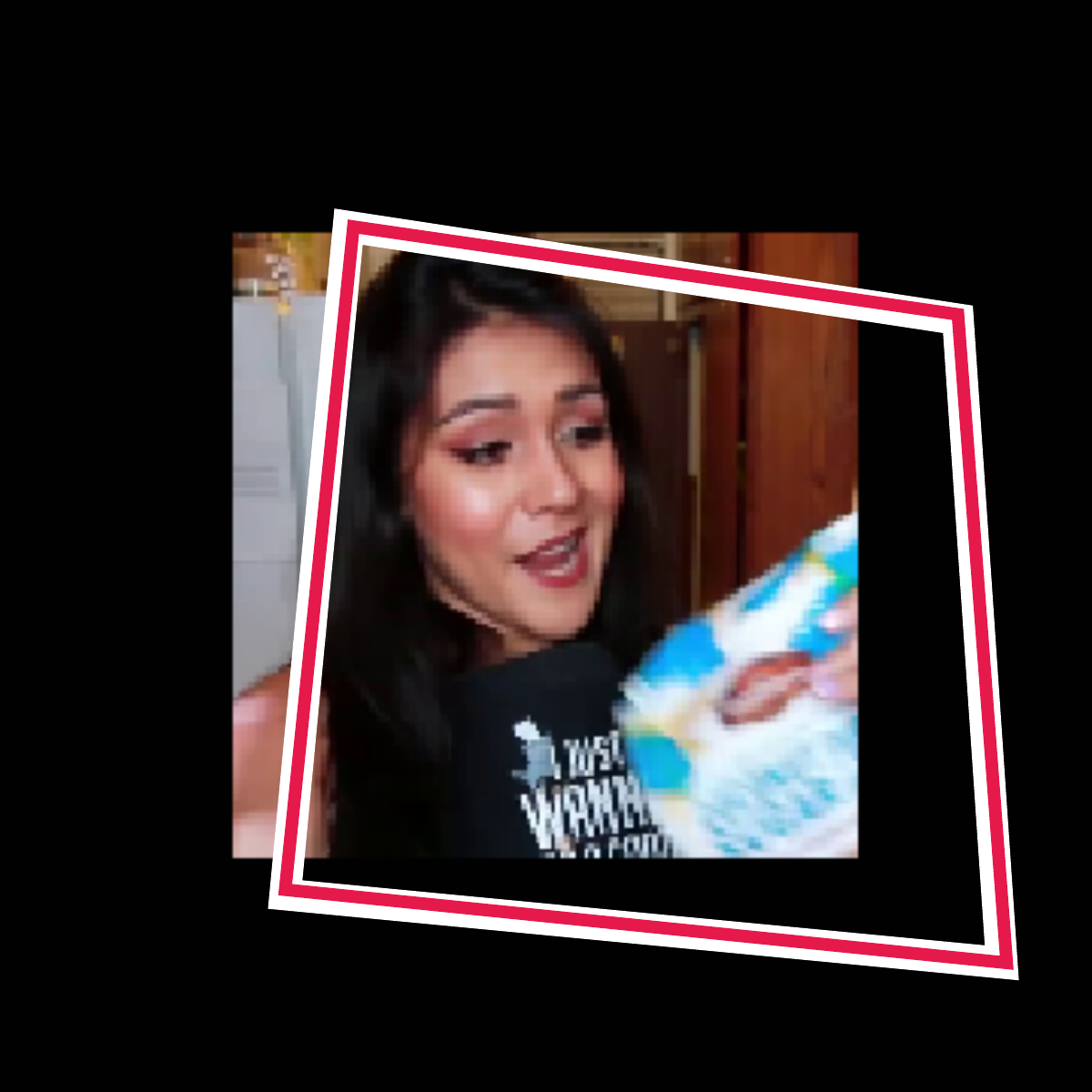}
    \vspace{-0.1mm}
    \\
    \hspace{-0.2cm}\small{SIFT+RANSAC~\cite{sift}} & 
    \hspace{-0.35cm}\small{DHN~\cite{detone2016deep}} & 
    \hspace{-0.35cm}\small{Our results}
\end{tabular}
\end{center}
\vspace{-0.1in}
\caption{Challenging examples for homography estimation.  Each figure shows one of the two input images. Each white box indicates where the other image in the pair will be warped to in the first image according to the ground truth homography and each red box indicates the result using the estimated homography.  As shown in the top row, the lack of texture fails the traditional SIFT + RANSAC method while the deep learning methods work well.  The example at the bottom is difficult as the moving person dominates the image. Compared to both the SIFT-based method and the deep homography estimation method~\cite{detone2016deep} , our dynamics-aware method can better estimate homography for a dynamic scene. 
\label{fig:figure1}}\vspace{-0.2in}
\end{figure}   

There are two categories of methods for homography estimation~\cite{szeliski2007image}. Direct photometric-based approaches search for an optimal homography that minimizes the alignment error between two input images. Sparse feature-based approaches first estimate sparse feature correspondences between the input images and then compute the homography from them. The direct approaches are robust to images with large textureless regions but are challenged by large motion. In contrast, while sparse feature-based approaches are more robust to large motion, they heavily depend on the quality of feature correspondences, which are often difficult to obtain in textureless regions or blurry images.

Recent deep homography estimation methods train a convolutional neural network to compute a homography between two images~\cite{chang2017clkn,detone2016deep,erlik2017homography,nguyen2018unsupervised}. These deep neural network methods leverage both local and global features and can often perform better than the traditional methods on challenging scenarios, such as images lacking texture, as shown in the first example of Fig.~\ref{fig:figure1}. However, how they handle challenging scenarios such as dynamic scenes has not been studied, as existing methods focus on image pairs that can be fully aligned using homographies. In practice, dynamic objects are common. Traditional non-deep learning approaches use a robust estimation algorithm like RANSAC~\cite{fischler1981random} to exclude them as outliers. We need to empower deep learning methods with the capability of being resistant to dynamic content to make them more practical. 

This paper investigates the problem of developing a deep learning method for homography estimation that is able to handle dynamic scenes without an iterative process like RANSAC. We first introduce a multi-scale deep convolutional
neural network to handle image pairs with a large global motion. This network first estimates the homography from
the low-resolution version of the input image pair and then
progressively refines it at the increasingly higher resolutions.
The architecture of the base neural network at each stage follows a VGG-style
network similar to existing deep homography estimation
methods~\cite{detone2016deep,nguyen2018unsupervised}. Our study shows that this multi-scale neural network can handle not only a large global motion, but also dynamic scenes to some extent when properly trained. 

To address the problem of homography estimation of a dynamic scene in a more principled way, we need to detect dynamic content and eliminate their effect on homography estimation. Actually, homography estimation and dynamic content detection are two tightly coupled tasks. According to the research on multi-task learning, training a neural network to simultaneously perform these two tasks can greatly improve its performance for both tasks. Therefore, we enhance our multi-scale network such that it jointly estimates the dynamics mask and homography for an image pair. To train this network, in addition to the homography loss, we use an auxiliary loss function that compares the dynamics mask from the ground-truth dynamics map that is estimated from the training data. This multi-task learning training strategy empowers our multi-scale homography estimation network to robustly handle dynamic scenes.

As there are no publicly available large number of image pairs or videos of scene dynamics that come with known ground-truth homographies, we collected 32,385 static video clips from YouTube with a Creative Commons license. We then applied random homography sequences to these static clips to produce the training examples. These examples contain a wide variety of dynamic scenes. As shown in our experiments, our neural networks trained on this synthetic dataset generalize well to real-world videos. 

This paper contributes to the research on homography estimation by investigating ways to develop and train deep neural networks that are robust against dynamic scenes.
First, we build a large video dataset with dynamic scenes for training deep homography estimation neural networks. Second, we develop a multi-scale neural network that can handle large motion and show that by carefully training, it can already reasonably accommodate dynamic scenes. Third, we develop a dynamics-aware homography estimation network by integrating a dynamics mask network into our multi-scale network for simultaneous homography and dynamics estimation. Our experiments show that our method can handle various challenges, such as dynamic scenes, blurriness, lack of texture, and poor lighting conditions.

\section{Related Work}
Homography estimation is one of the basic computer vision problems~\cite{Hartley2003-ca}. According to multi-view geometry theory,
two images of a planar scene or taken by a rotational
camera can be related by a $3 \times 3$ homography matrix \textbf{H}:

\vspace{-0.18in}
\begin{equation} \label{eqn:hmg}
    \hat{\textbf{x}} = \textbf{H}\textbf{x}
\vspace{-0.12in}
\end{equation}

\noindent where $\hat{\textbf{x}}$ and $\textbf{x}$ are the homogeneous coordinates of two corresponding points in the two images. Note, the above equation is only valid for corresponding points on static objects. Below we first briefly describe two categories of off-the-shelf algorithms for homography estimation and then discuss the recent deep neural network based approaches. 

Pixel-based approaches directly search for an optimal homography matrix that minimizes the alignment error between two input images. Various error metrics between two images and parameter searching algorithms, such as hierarchical estimation and Fourier alignment, have been developed to make direct approaches robust and efficient. These direct methods 
are robust to images lacking in texture, but often have difficulty in dealing with large motion. Feature-based approaches are now popular for homography estimation. They
first estimate local feature points using algorithms, such as
SIFT~\cite{sift} and SURF~\cite{surf}, and then match feature points
between two images. For a video, corner points are often detected and then tracked
across two consecutive frames for efficiency~\cite{klt}. Given the set of
corresponding feature pairs, an optimal homography matrix
can be obtained by solving a least squares problem
based on Eq.~\ref{eqn:hmg}. In practice, errors can occur during
feature matching and feature points can come from moving
objects. Therefore, a robust estimation algorithm like
RANSAC~\cite{fischler1981random} and Magsac~\cite{barath2019magsac} is often used to remove the outliers. The
performance of feature-based approaches depends on local
feature detection and matching. For images suffering from
blurriness or lacking in texture, they often cannot work well.

\begin{figure*}[t]
\begin{center}
\begin{tabular}{c}    
    \hspace{-0.2cm}\includegraphics[width=1.0\linewidth]{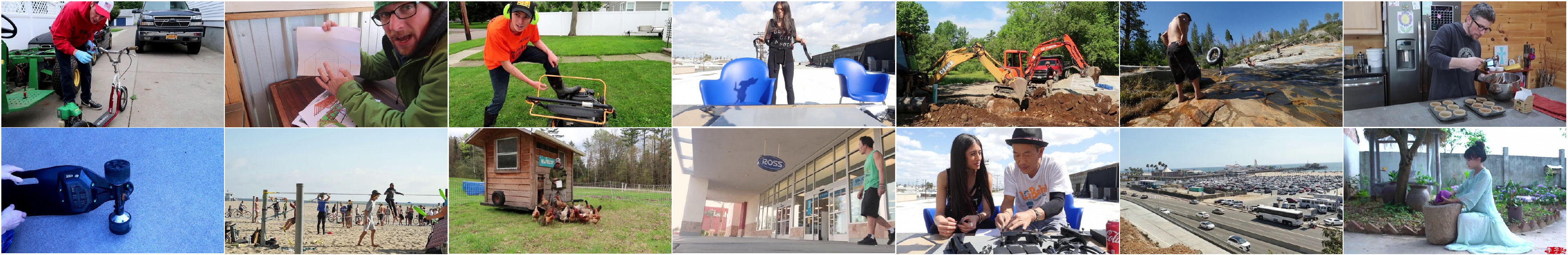}
\end{tabular}
\vspace{-0.65cm}
\end{center}
\caption{Static video samples in our dynamic-scene homography dataset.~\label{fig:dataset:examples}}\vspace{-0.25in}
\end{figure*}

Our work is more related to the recent deep learning approaches to homography estimation. In their seminal work, DeTone \etal developed VGG-style deep convolutional
neural networks for homography estimation. They showed that a deep neural network can effectively compute the homography between two images~\cite{detone2016deep}. Nguyen~\etal extended 
this work by training the neural network using a photometric loss that measures the pixel error between a warped input image and the other one. This photometric loss allows for the unsupervised training of the neural network without ground-truth homographies~\cite{nguyen2018unsupervised}. To deal with the large motion between two images, Nowruzi \etal developed a hierarchical neural network that iteratively refines homography estimation~\cite{erlik2017homography}. Chang \etal designed a Lucas-Kanade
layer that is able to regress the homography between two
images from their corresponding feature maps extracted by
a shared convolutional neural network~\cite{chang2017clkn}. This Lucas-Kanade network can be cascaded to progressively refine the estimated homography. Zeng~\etal developed a perspective field neural network to model the pixel-to-pixel bijection between two images and used that for homography estimation~\cite{zeng2018rethinking}.  These deep homography estimation approaches are shown successful on images of a static scene; however, they do not consider dynamic scenes.

Our multi-scale neural network extends the multi-stage
approaches discussed above. Compared to Nowruzi~\etal\cite{erlik2017homography}, our method starts from low-resolution versions of
the input images and gradually increases the input image
sizes, instead of taking the original input images as input at
each stage. This makes our method more robust to large motion.
Compared to Chang~\etal\cite{chang2017clkn}, our method pre-aligns
the input images to the next stage using the homography
estimated in the previous stage to minimize the global motion.
This helps the late-stage network to account for the
global motion. More importantly, we further enhance our
multi-scale neural network with a dynamics mask network to handle
dynamic scenes, which was not considered in the previous
neural network-based methods.

\section{Homography Dataset of Dynamic Scenes}
\label{sec:data}
Existing deep neural networks for homography estimation are trained with image pairs that can be perfectly aligned. A common approach is to use a subset of the MS-COCO image set~\cite{Lin2014-or}, and warp each image using a known homography to form an image pair~\cite{chang2017clkn,detone2016deep,erlik2017homography}. Since we aim to train a neural network that can handle dynamic scenes, we cannot produce a dataset in this way.

Ideally, each image pair in our dataset should contain dynamic scenes with a known homography. To our best knowledge, there is not such a public homography dataset
that is large enough for training a deep neural network. Our solution is to first collect a large set of videos capturing dynamic scenes while the cameras are held static, and then apply known homography sequences to them. Specifically, we downloaded $877$ videos with a Creative Commons License from YouTube. From these videos, we extracted $32,385$ static video clips and then applied a known homography sequence to each of them to generate image/video pairs. Fig.~\ref{fig:dataset:examples} shows some sample frames from this video set. 

\noindent\textbf{Static video clip detection.} Since our goal is to build a reliable homography benchmark to train a deep neural network, it is critical that each image pair in our dataset can be perfectly aligned by a homography (except moving objects.) Thus, instead of using a structure-from-motion method to estimate and analyze camera motion, we use a more conservative approach \emph{aiming for a very high precision rate at the cost of a low recall rate when identifying static clips}.

We observe that if a video clip has camera motion, its boundary area changes temporally. Also, the central part of a video clip may change significantly due to the scene motion if the video is captured by a static camera. Accordingly, for every two consecutive frames, we calculate the similarity between their boundary areas. Specifically, we first resize each video frame to $256\times256$. We consider the outer boundary with a $5$-pixel thickness. We subdivide the boundary area into $32$ boundary blocks of size $32\times5$ pixels. If more than $25\%$ of blocks stay the same between two consecutive frames, we consider that there is no camera motion between the two frames. We consider a boundary block unchanged over time if more than $90\%$ of its pixels only slightly change the color ($\delta_c \leq 6.67$). 
To further remove non-static video clips, we estimate optical flows between frames in each video clip using the PWC-Net algorithm~\cite{Sun2019-im}. Since the difference between two consecutive frames is small, we skip $\delta t=7$ frames to compute the optical flow. We remove the clips if the areas of moving pixels in the frames are greater than $65\%$. Finally, we consider a video clip static if every two consecutive frames are static. To avoid the drifting error, we also enforce that the following frames in a sequence must also be considered static with regard to the first frame of the sequence. We extract static video clips that have a minimum length of $10$ frames from a video using a greedy search method. 

We finally manually examine the resulting video clips to remove those misidentified as static ones. In total, we obtained $32,385$ static video clips. The average length of these video clips is $22$ frames and the scene motion ranges from 0-25 pixels. We split the dataset into three portions, $70\%$ for training, $20\%$ for testing and $10\%$ for validation.

\noindent\textbf{Image pair generation.}
Given a sequence of $n$ frames $\{I_i, 1\leq i \leq n\}$ in a video clip, we randomly sample 2 frames $I_j$ and $I_k$, with $1 \leq j,k \leq n$, and $|j-k|\leq5$. We then use a similar method to Detone~\etal\cite{detone2016deep} to generate a pair of image patches, one from $I_j$ and the other from $I_k$. Specifically, we first randomly crop an image patch $I^p_j$ of size $128\times128$ at location $[x_p,y_p]$ in $I_j$. We then randomly perturb the coordinates of the four corner points of $I^p_j$ by a value $r, -32\leq r \leq 32$. We use the four corner displacements to compute the corresponding homography $\mathbf{H}$. Then, we apply the inverse of this homography,  $\mathbf{H}^{-1}$, on the other image $I_k$: $\hat{I}_k=\mathbf{H}^{-1}(I_k)$, and then extract the corresponding patch at the same location $[x_p,y_p]$ to obtain image patch $\hat{I}^p_k$. Each training sample contains $I^p_j$, $\hat{I}^p_k$, and their corresponding ground-truth homography $\mathbf{H}$.
 
\begin{figure}
\begin{center}
\begin{tabular}{c}    
    \hspace{-0.3cm}\includegraphics[width=1.01\linewidth]{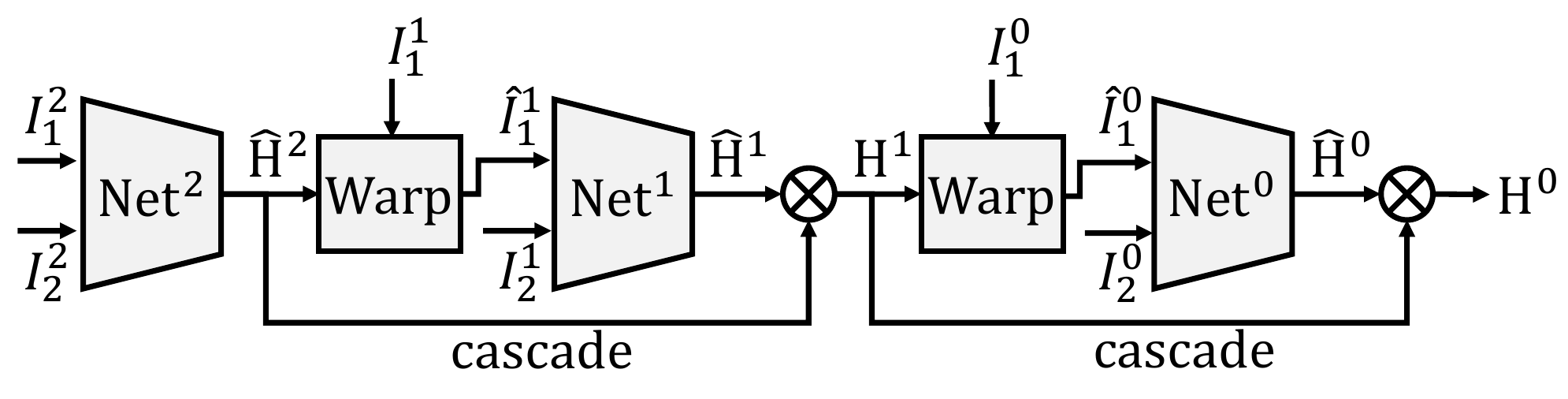}
\end{tabular}
\end{center}\vspace{-0.2in}
\caption{Multi-scale neural network for homography estimation.
This network progressively estimates and refines homography
from coarse to fine.
\label{fig:method:multi-scale}}
\vspace{-0.23in}
\end{figure}   

\vspace{-0.05in}
\section{Homography Estimation Neural Networks}
\vspace{-0.05in}
We first introduce a multi-scale deep homography estimation
network that handles large motion, and then describe
how we improve it with a dynamics mask network to better handle dynamic scenes.

\vspace{-0.05in}
\subsection{Multi-scale Neural Network}
\vspace{-0.05in}
\label{sec:multi}

Our neural network takes two gray-scale images of size $128 \times 128$ as input, and outputs the homography between them. Following the previous work~\cite{detone2016deep,erlik2017homography,nguyen2018unsupervised}, we use the displacements of the four image corners to represent the homography. As reported in previous work~\cite{baker2004lucas,chang2017clkn,erlik2017homography},
progressively estimating and refining a homography with a multi-stage procedure is helpful to cope with large global motion between two images. We extend these methods and design a multi-scale multi-stage neural network to estimate the homography for a pair of images from coarse to fine.

Fig.~\ref{fig:method:multi-scale} illustrates a three-scale version of our multi-scale
homography neural network. Given a pair of images ($I_1$, $I_2$), we build a pair of image pyramid ($I^k_1$, $I^k_2$), where $k$ indicates the pyramid level. The down-scaling factor at level $k$ is $2^k$. We start from the highest pyramid
level ($I^2_1$, $I^2_2$) and use the base network $\mbox{Net}^2$ to estimate a homography $\hat{\mathbf{H}}^2$. Then, we warp $I^1_1$ using $\hat{\mathbf{H}}^2$ to obtain a pre-aligned image pair ($\hat{I}^1_1$ , $I^1_2$). The homography of the unaligned image pair ($I^1_1$, $I^1_2$) can be computed by cascading $\hat{\mathbf{H}}^2$ and $\hat{\mathbf{H}}^1$.

\vspace{-0.17in}
\begin{equation}\label{eqn:cascade}
    \mathbf{H}^1 = \hat{\mathbf{H}}^1 \mathbf{S}^{-1} \hat{\mathbf{H}}^2\mathbf{S}
\vspace{-0.12in}
\end{equation}
where $\mathbf{H}^1$ is the homography of the unaligned image pair
($I^1_1$, $I^1_2$) and $\mathbf{S}$ is a scaling matrix that down-samples an image
by half to account for the different image sizes at the two scales. We continue by using $\mathbf{H}^1$ to pre-align ($I^0_1$ , $I^0_2$) as input for the subsequent base network and obtain the homography $\hat{\mathbf{H}}^0$ and compute the homography for the original image pair in a similar way to Eq.~\ref{eqn:cascade}.

\begin{figure}
\begin{center}
\begin{tabular}{c}    
    \hspace{-0.0cm}\includegraphics[width=1.01\linewidth]{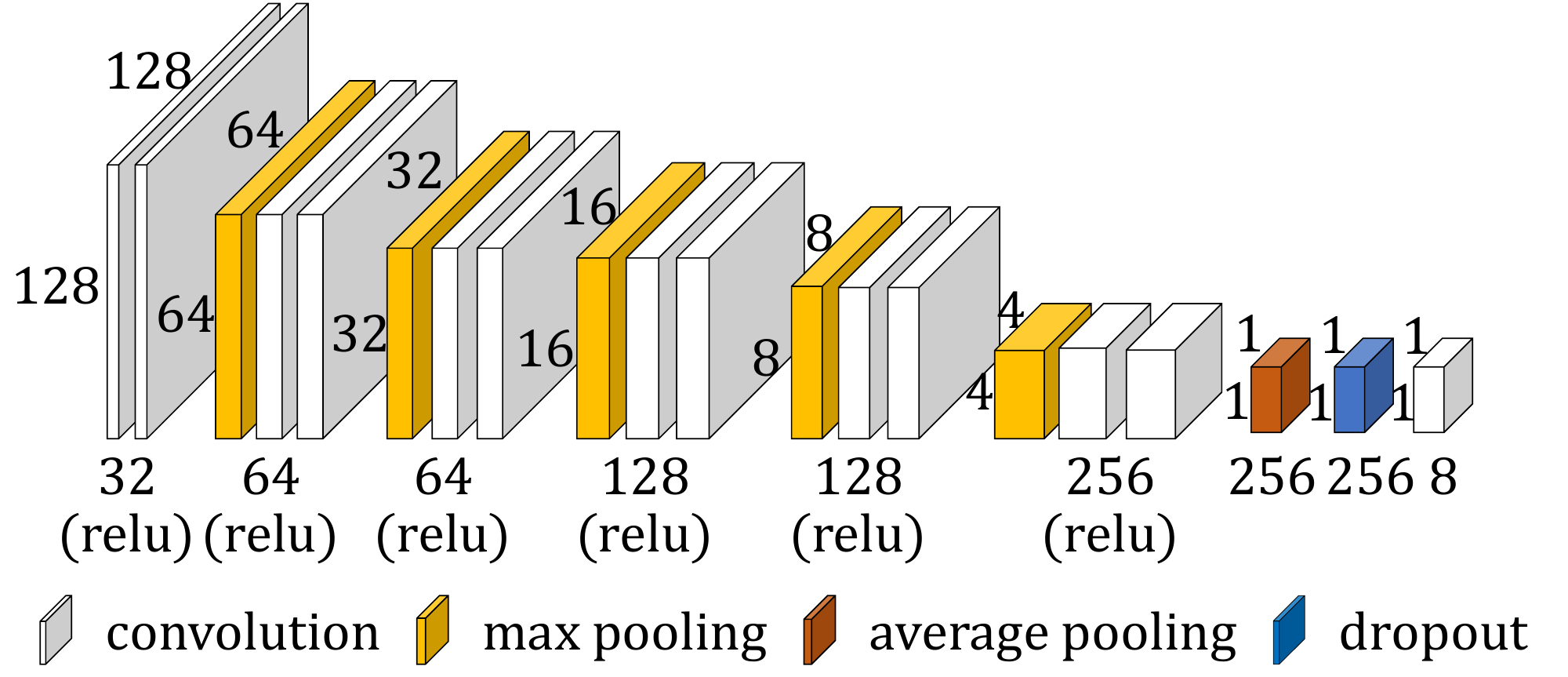}
\end{tabular}
\end{center}
\vspace{-0.17in}
\caption{Architecture of the base network Net$^0$.\label{fig:method:net_archi_base_net0}}
\vspace{-0.25in}
\end{figure}   

\begin{figure*}
\begin{center}
\begin{tabular}{c}    
    \includegraphics[width=1.0\linewidth]{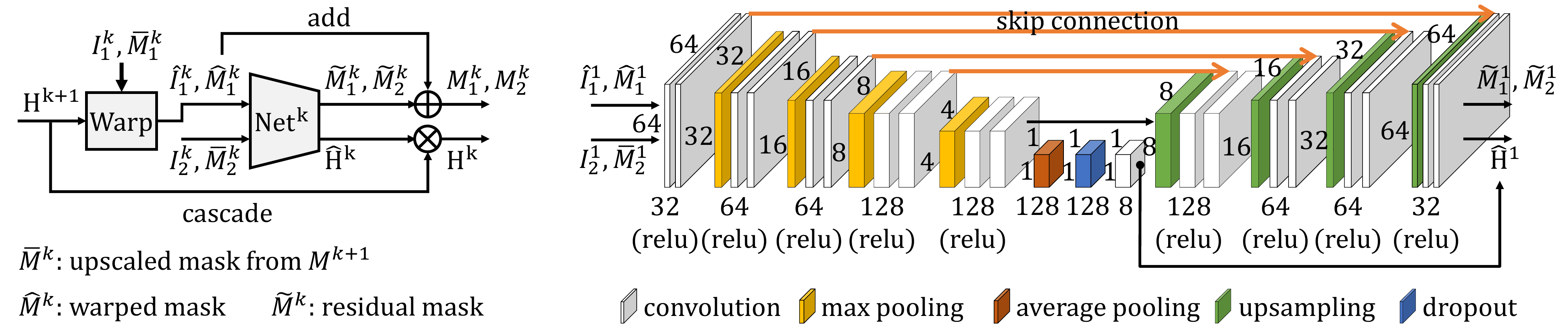}
    \\
\end{tabular}
\end{center}
\vspace{-0.15in}
\caption{Mask-augmented Deep Neural Network at level $k=1$\label{fig:method:multi-scale-mask}. The network is augmented with a convolutional decoder that helps predict areas of dynamic objects to improve the homography prediction.}
 \vspace{-0.2in}
\end{figure*}  

We use a variation of the neural network introduced by DeTone~\etal\cite{detone2016deep} as the base network at each stage. Fig.~\ref{fig:method:net_archi_base_net0} shows the architecture of the base network Net$^0$. It starts with twelve $3 \times 3$ fully convolutional layers, each coupled with Batch Normalization~\cite{ioffe2015batch}  and a Rectified Linear Unit layer (ReLU). Every two consecutive convolutional layers are followed by a max pooling layer. These convolutional layers are followed by an average pooling layer, a dropout layer and a $1 \times 1$ convolution layer that outputs a homography vector of length 8. The architecture of other base networks Net$^k$ is similar except that each has $2k$ less convolutional layers due to the smaller input image size.

To train this multi-scale neural network, we first calculate the $l_2$ loss between the estimated and ground-truth homography, each parameterized using a corner displacement vector. We then compute the total loss at all the scales as the loss function. 

When our multi-scale neural network is trained using the examples derived from the MS-COCO dataset in a similar way to previous work~\cite{detone2016deep}, it works well with image pairs with only static background, for which the transformation can
be perfectly modeled by a homography. However, when it is used to estimate homographies for image pairs with scene motion, the results are less accurate as reported in Sec.~\ref{sec:exp}. This is expected as the training examples derived from the MS-COCO dataset do not contain dynamic scenes. 

To further examine the capability of this multi-scale neural network, we trained another version on our dynamic dataset described in Sec.~\ref{sec:data}. We found that training this multi-scale network on our dataset can improve its capability in handling dynamic scenes, as found in our experiments (Sec.~\ref{sec:exp:motion}). To better handle scene motion, we develop a mask-augmented network as described in the next section.

\subsection{Mask-augmented Deep Neural Network}
\label{sec:mask}

To better handle dynamic scenes, we need to identify dynamic content in the scene. Actually, dynamic content detection and homography estimation are two tightly coupled tasks. An accurate estimation of homography helps robustly detect dynamic content while correctly identifying dynamic content helps with accurate homography estimation. Based on the research on multi-task learning, jointly training a model to perform these two tasks simultaneously can enable the success of both tasks~\cite{Caruana:1993,ruder2017overview}. Accordingly, we improve our base homography network so that it estimates both the dynamics maps and the homography for a pair of images. To this end, we incorporate a dynamics mask estimator into our base homography neural network, as shown in the right of Fig.~\ref{fig:method:multi-scale-mask}. This new network uses the same architecture as the previous base network  except that we add a sub-neural network to regress a pair of dynamics maps. We refer to this new base network as a mask-augmented base network. 

We embed this mask-augmented base network into our multi-scale network in a similar way as before. As shown in the left of Fig.~\ref{fig:method:multi-scale-mask}, the mask-augmented base network Net$^{k+1}$ outputs a pair of dynamics maps $(M_1^{k+1}, M_2^{k+1})$ and a homography $\hat{\mathbf{H}}^{k+1}$, from which we can compute a cascaded homography $\mathbf{H}^{k+1}$ according to Eq.~\ref{eqn:cascade}. Then, we first upsample the pairs of dynamics maps to match the image size in the next level $k$ and obtain $(\bar{M}_1^k, \bar{M}_2^k)$. Then we warp $\bar{M}_1^k$ using the homography $\mathbf{H}^{k+1}$ and obtain $\hat{M}_1^k$ to match the warped input image $\hat{I}_1^k$, and finally concatenate $(\hat{M}_1^k, \bar{M}_2^k)$ and ($\hat{I}^k_1$, $I^k_2$) as input for the new base network at the next level $k$. We further improve the performance of this mask-augmented network by first estimating the mask residuals $(\tilde{M}_1^k, \tilde{M}_2^k)$ and then obtaining the final masks by adding the residuals to the input upsampled masks $(\hat{M}_1^k, \bar{M}_2^k)$.

This mask-augmented homography neural network is difficult to train as it cannot  automatically figure out the role of the dynamics maps. According to multi-task learning~\cite{Caruana:1993,ruder2017overview}, having an extra supervising signal helps train a neural network. So we estimate the ground-truth dynamics maps and use them as the extra supervising signals. Specifically, since each image pair $(I_i, I_j)$ is created from two frames of a static video, we first estimate the optical flow between the two video frames using PWC-Net~\cite{Sun2019-im} and then create a ground truth mask by labelling a pixel as 1 if the flow magnitude is greater than 1 pixel and 0 otherwise. We then generate a pair of masks $(M_i, M_j)$, one for each image in the pair, by cropping and warping the ground truth mask in the same way as we generate the image pair $(I_i, I_j)$.

Now we have two loss functions: a homography loss $l_f$ described in Sec.~\ref{sec:multi} and a dynamics mask-based loss $l_d$. We calculate $l_d$ as a binary cross entropy loss as follows~\cite{bishop2006pattern}.

{\small{
\vspace{-0.2in}
\begin{equation}
    l_d=\frac{-M_{g}\cdot \log(M_{p})^T-(1-M_{g})\cdot \log(1-M_{p})^T}{|M_{g}|}
\vspace{-0.07in}
\end{equation}
}}where $M_{p}$ and $M_{g}$ are the predicted dynamics mask and the ground-truth mask, each arranged as a row vector. $|M_{g}|$ is the total number of pixels in $M_{g}$. We combine the two losses to train our network: $L = \sum_k{\sigma_f l_f^k + \sigma_d l_d^k}$ where $l_f^k$ and  $l_d^k$ are the homography loss and mask loss at scale $k$ and $\sigma_f$ and $\sigma_d$ are weights.

\begin{figure*}[t]
\begin{center}
\begin{tabular}{cc}    
    \includegraphics[width=0.48\linewidth]{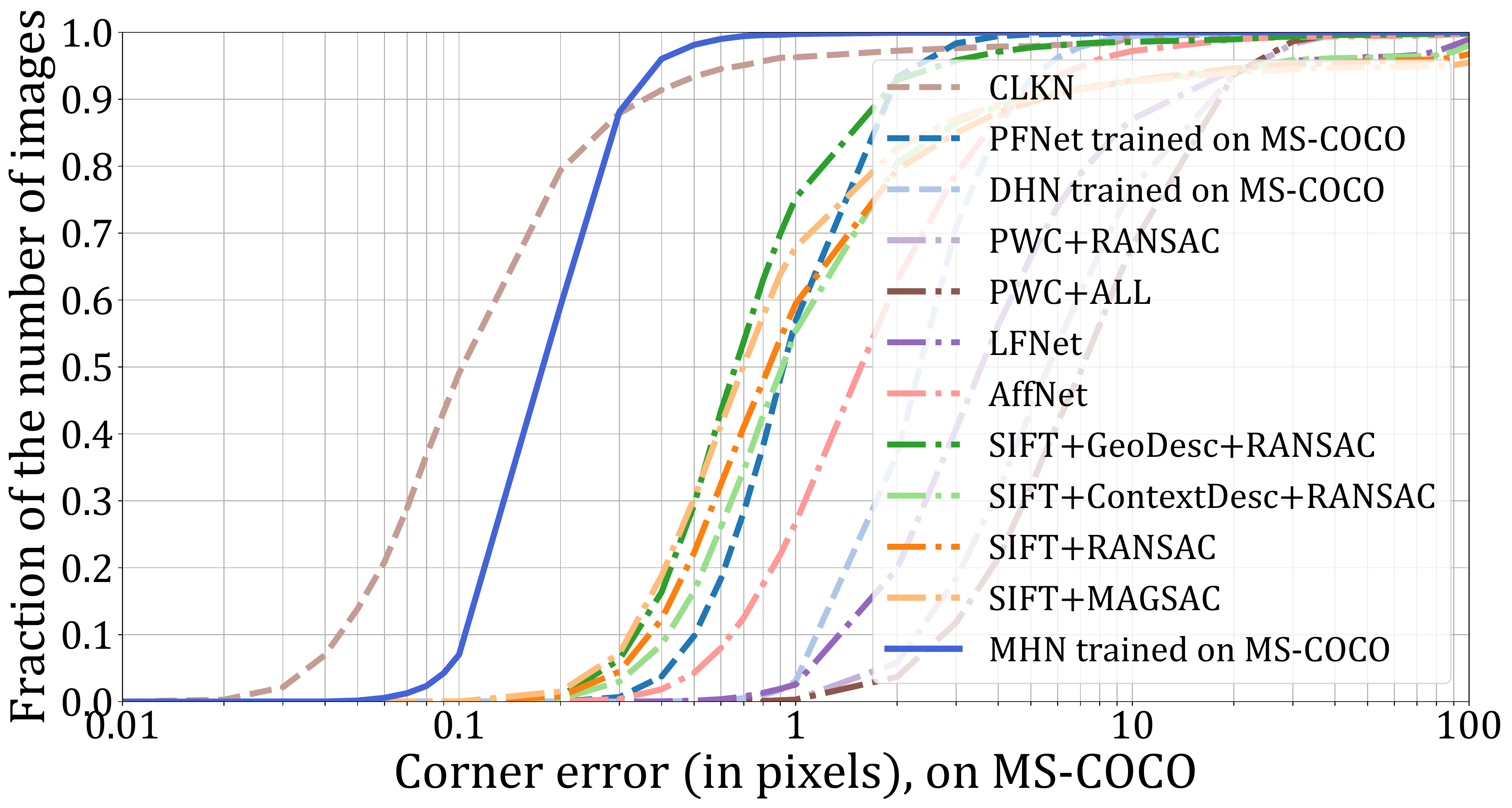}
    &
    \includegraphics[width=0.48\linewidth]{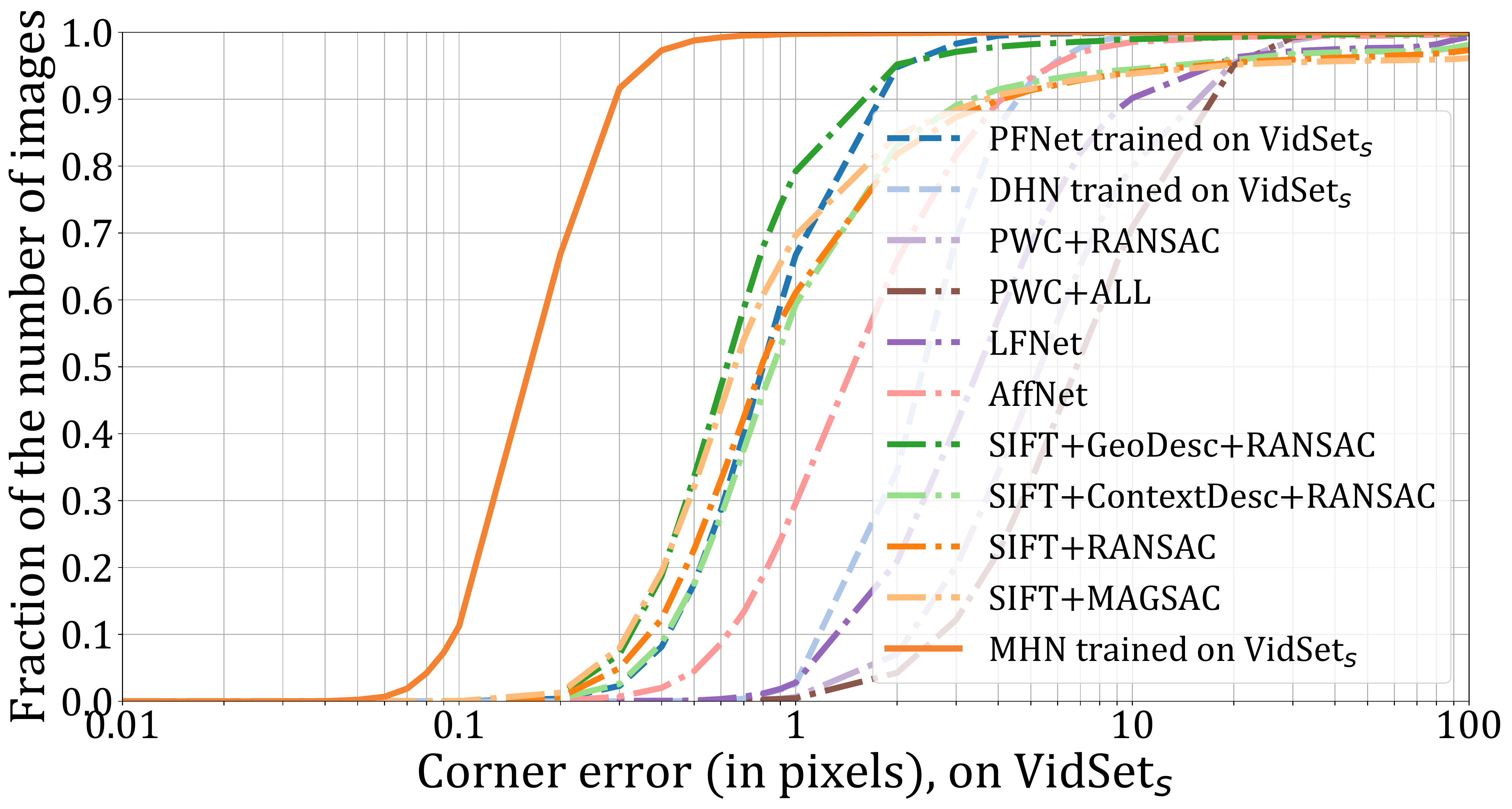}
    \\
    \vspace{-0.25cm}
\end{tabular}
\vspace{-0.33in}
\end{center}
 \caption{Evaluation on static scenes.
\label{fig:exp:compare_sota_static}}\vspace{-0.25in}
\end{figure*}    

\noindent\textbf{Implementation details.} Our neural networks are implemented using TensorFlow. We initialize the weight values using the Xavier algorithm~\cite{glorot2010understanding} and the biases with zero. We use the Adam optimization algorithm to train the neural networks~\cite{kingma2014adam} using an exponentially decaying learning rate with an initial value of $10^{-4}$, a decay step of $10^5$, and a decay rate of $0.96$. The dropout keep probability is set to $0.8$. The mini-batch size is $32$. 

We train the mask-augmented network in three steps. First, we train it on our dynamic-scene dataset with $\sigma_f=1$ and $\sigma_d=0$ for $2\times10^6$ global iterations so that the network learns to predict homography only with ground-truth homography labels. We continue training it for another $10^6$  iterations with $\sigma_f=1$ and $\sigma_d=10$ so that it learns to predict both the homography and dynamics masks by leveraging the ground-truth masks. Finally, we train it for another $10^6$  iterations with $\sigma_f=1$ and $\sigma_d=0$ so that it learns to leverage the predicted dynamics masks to ultimately boost the performance of homography estimation for dynamic scenes.

\section{Experiments}
\label{sec:exp}
 We evaluate two versions of our method: the multi-scale homography network (MHN) and the dynamics mask-augmented network (MHN$_m$). We compare them to both feature matching-based and deep learning-based methods. The first category of methods use the SIFT feature or its recent variations followed by a robust estimation method, including SIFT+RANSAC~\cite{sift}, SIFT+MAGSAC~\cite{barath2019magsac}, SIFT+GeoDesc+RANSAC~\cite{luo2018geodesc}, SIFT+ContextDesc+RANSAC~\cite{luo2019contextdesc}, LF-Net~\cite{Ono2018-dr}, and AFFNET~\cite{mishkin2018repeatability}. We also use PWC-Net~\cite{Sun2019-im}, a state-of-the-art optical flow method, to estimate dense correspondences between input images and then use RANSAC to estimate the homography. The deep learning homography estimation methods include CLKN~\cite{chang2017clkn}, DHN~\cite{detone2016deep}, and PFNet~\cite{zeng2018rethinking}.

Our experiments use three datasets. The first is derived from the MS-COCO~\cite{Lin2014-or}, following the procedure in the recent  work~\cite{chang2017clkn,detone2016deep}. The image size is $128\times128$  and the image corners are randomly shifted in the range of [-32, 32] pixels. Please refer to~\cite{chang2017clkn,detone2016deep} for more details. The second, named as VidSet$_s$, is the static version of our dynamic-scene dataset by creating image pairs from a single video frame so that there is no dynamic content in each image pair. The third is our dynamic-scene dataset described in Sec.~\ref{sec:data}, named as VidSet$_d$. Following~\cite{chang2017clkn}, we use the mean corner error as the evaluation metric: $e_c=\frac{1}{4}\sum_{j=1}^4||c_j-\hat{c}_j||_2$
where $c_j$ and $\hat{c}_j$ are corner $j$ transformed by the estimated homography and the ground-truth homography, respectively.

\subsection{Evaluation on Static Scenes}
\label{sec:exp:static}

\begin{figure}
\begin{center}
\begin{tabular}{c}    
    \hspace{-0.1cm}\includegraphics[width=1.0\linewidth]{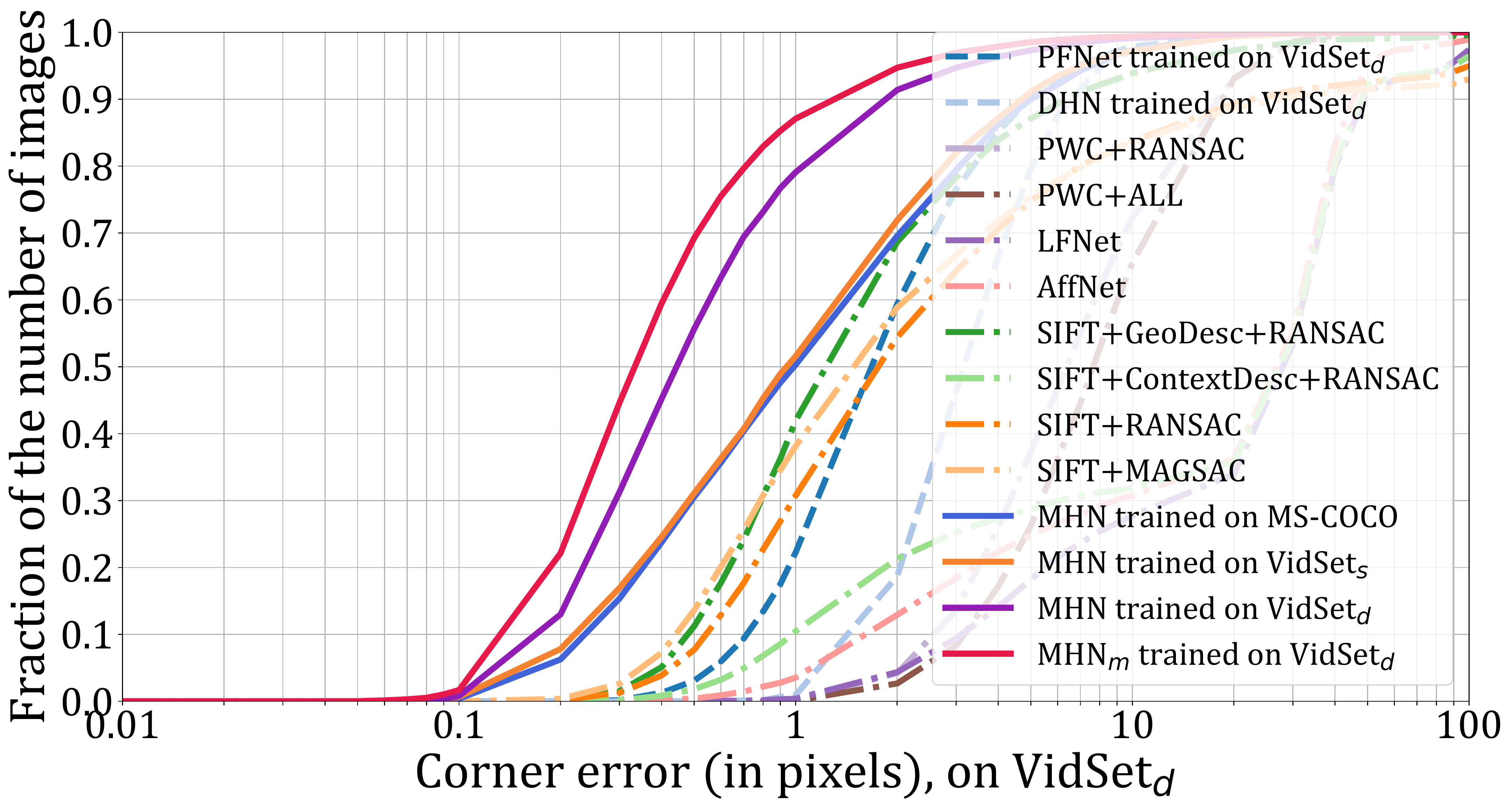}
\end{tabular}
\vspace{-0.65cm}
\end{center}
\caption{Evaluation on dynamic scenes.
\label{fig:exp:compare_sota_ours_dynamic}}
\vspace{-0.6cm}
\end{figure}   

Existing deep homography estimation methods focus on pairs of images that can be fully aligned by homography. We compare our multi-scale network (MHN) with these methods on both MS-COCO and VidSet$_s$. As shown in Fig.~\ref{fig:exp:compare_sota_static}, our multi-scale network described in Sec.~\ref{sec:multi} outperforms the competitive learning-based methods~\cite{detone2016deep} and~\cite{zeng2018rethinking} but performs slightly worse than CLKN~\cite{chang2017clkn} in the high-precision region. Moreover, our method outperforms all the competitive feature matching or flow-based methods. This is consistent with the reports from the previous work~\cite{chang2017clkn,detone2016deep} that deep neural network approaches can often perform better than the conventional matching-based methods.

\subsection{Evaluation on Dynamic Scenes}
\label{sec:exp:motion}

\begin{figure}
     \hspace{-0.2cm}
    \includegraphics[width=1.0\linewidth]{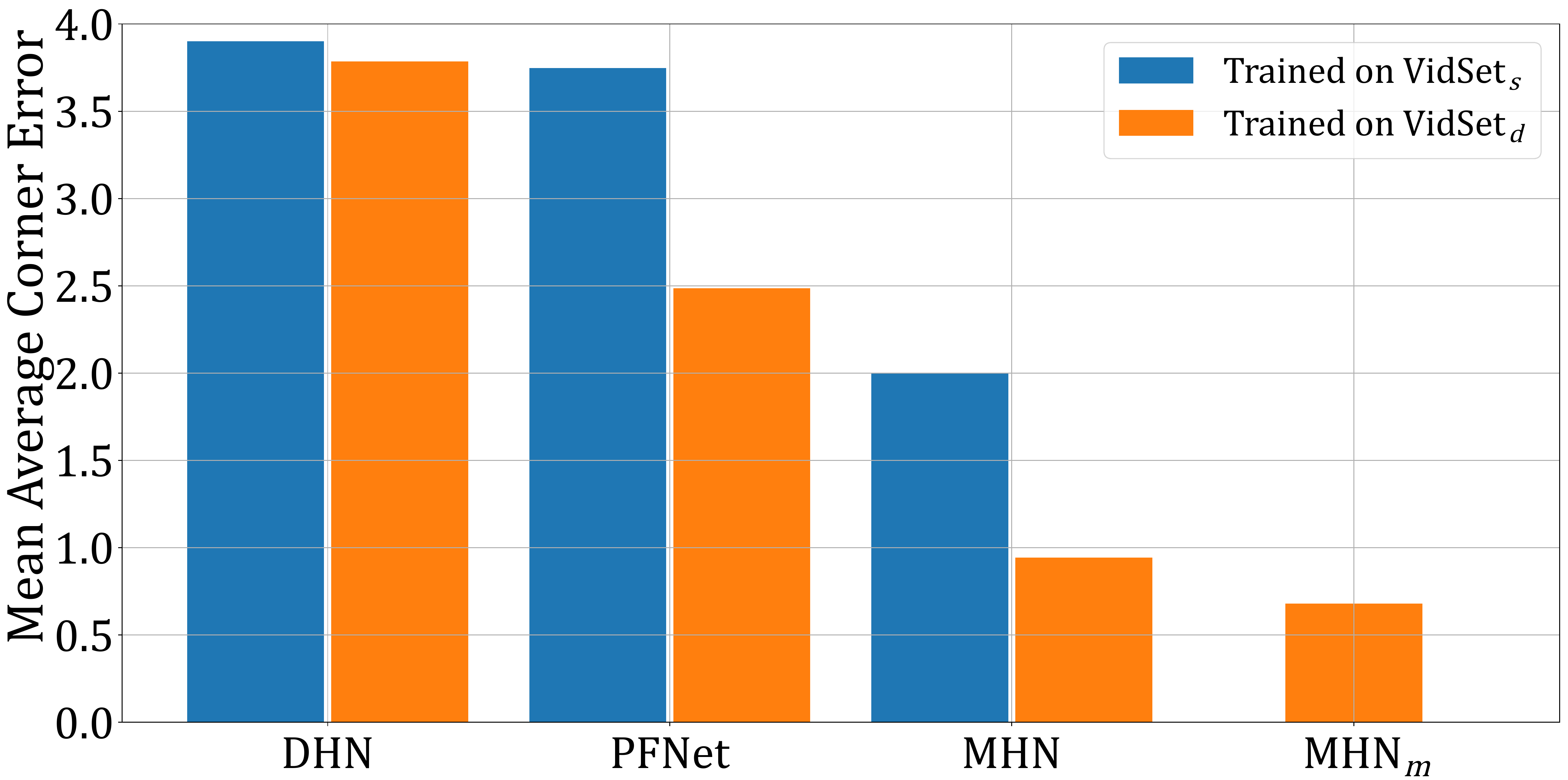}
\vspace{-0.1cm}
\caption{Effect of training datasets. Our dynamic-scene dataset VidSet$_d$ enables deep homography estimation networks better handle dynamic scenes than its static version VidSet$_s$.~\label{fig:exp:effect_of_dynamic_dataset}}
\vspace{-0.25in}
\end{figure}    

In this test, we train and test our networks MHN and MHN$_m$ and the other competitive networks, including DHN~\cite{detone2016deep} and PFNet~\cite{zeng2018rethinking}, on our dynamic-scene dataset VidSet$_d$. Fig.~\ref{fig:exp:compare_sota_ours_dynamic} shows that our networks significantly outperform the existing learning-based approaches and matching-based approaches in handling dynamic scenes, even though we train the existing deep learning methods on our dynamic-scene dataset. In addition, when trained on VidSet$_d$, our multi-scale network MHN can already handle dynamic scenes to some extent. On around 80\% of examples, it achieves a good accuracy of $\leq$1.0 pixel. Our dynamics mask augmented network MHN$_m$ further improves its performance, and achieves the same accuracy for more than 85\% of the testing mages. We were not able evaluate the performance of CLKN~\cite{chang2017clkn} on VidSet$_d$ since their code is not available. However, we also expect that our mask network and our VidSet$_d$ can also be married to CLKN to better handle dynamic scenes.

\noindent\textbf{Effect of training sets.} We trained two versions of each homography network, one on the static version of our dynamic-scene dataset (VidSet$_s$) and the other on our dynamic-scene dataset (VidSet$_d$). We  tested them on the testing set of VidSet$_d$. As shown in Fig.~\ref{fig:exp:effect_of_dynamic_dataset}, our dynamic-scene dataset can greatly improve most of these deep homography estimation networks in handling dynamic scenes.

\begin{figure}
 \hspace{-0.2cm}
      \includegraphics[width=1.0\linewidth]{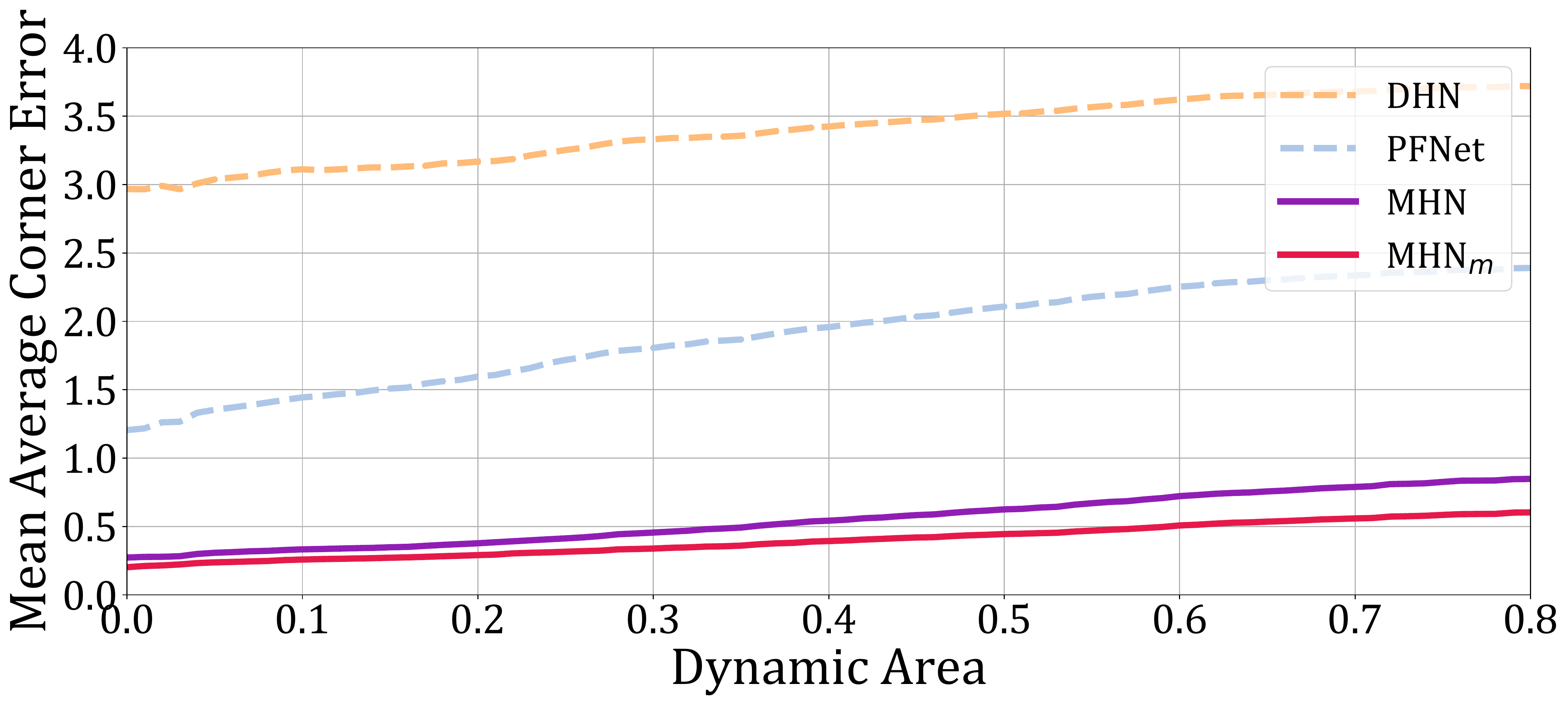}
      \vspace{-0.18cm}
    \caption{Effect of the size of dynamic area. Compared to existing deep homography estimation methods, such as DHN~\cite{detone2016deep} and PFNet~\cite{zeng2018rethinking}, our methods are robust against a large dynamic area.\label{fig:exp:effect_motion_area}}
\vspace{-0.24in}
\end{figure}

\noindent\textbf{Effect of the dynamic area size.} To examine how our methods work on images with a various amount of dynamic content, we calculate the dynamic area ratio of each example. Then, we create multiple versions of the testing set. For each set, we select a dynamic area ratio threshold and only include examples with the dynamic area ratio smaller than that threshold. We then test our methods on these testing sets. As reported in Fig.~\ref{fig:exp:effect_motion_area}, both of our methods, MHN and MHN$_m$ are more stable than the other deep homography methods when the dynamic area increases. 

\subsection{Discussions}

\begin{figure}
 \hspace{-0.2cm}
    \includegraphics[width=1.0\linewidth]{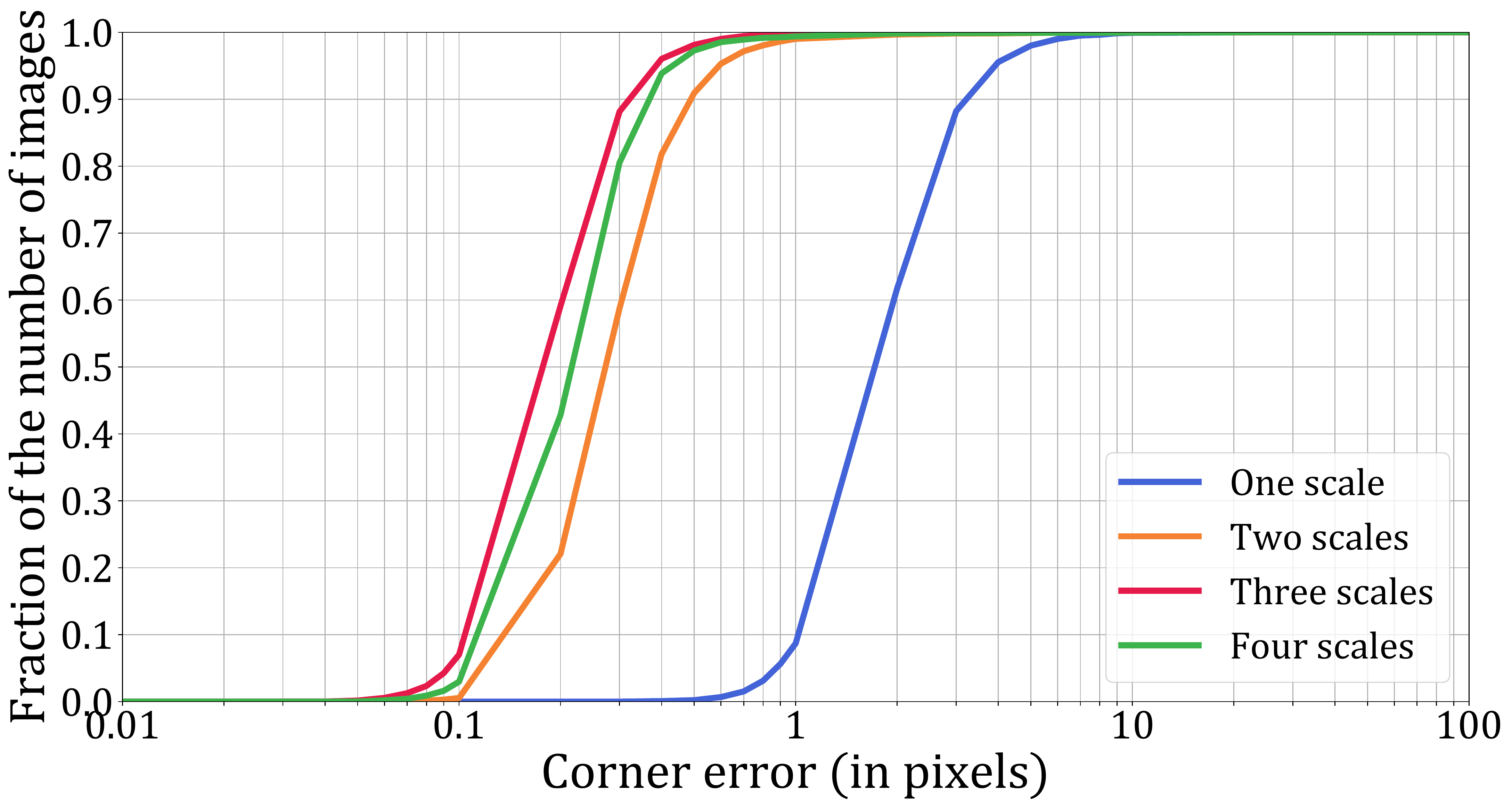} 
\vspace{-0.05in}
\caption{Effect of the number of scales in our multi-scale homography estimation network on MS-COCO.}\label{fig:exp:effect_of_different_number_of_scales}
 \vspace{-0.25in}
\end{figure}

\noindent\textbf{Scale selection.} An important hyper-parameter of our multi-scale neural network is the number of scales. We found that when increasing the number of scales from one to three, our network can be trained significantly faster. As shown in Fig.~\ref{fig:exp:effect_of_different_number_of_scales}, it can handle larger global motion. But when the number of scale goes beyond three, the training becomes unstable. We attribute this to the very small image size that is processed by the first base neural network. This is similar to what Chang~\etal reported~\cite{chang2017clkn}.

\begin{figure*}[t]
\begin{center}
\begin{tabular}{c}    
    \hspace{-0.0cm}\includegraphics[width=0.82\linewidth]{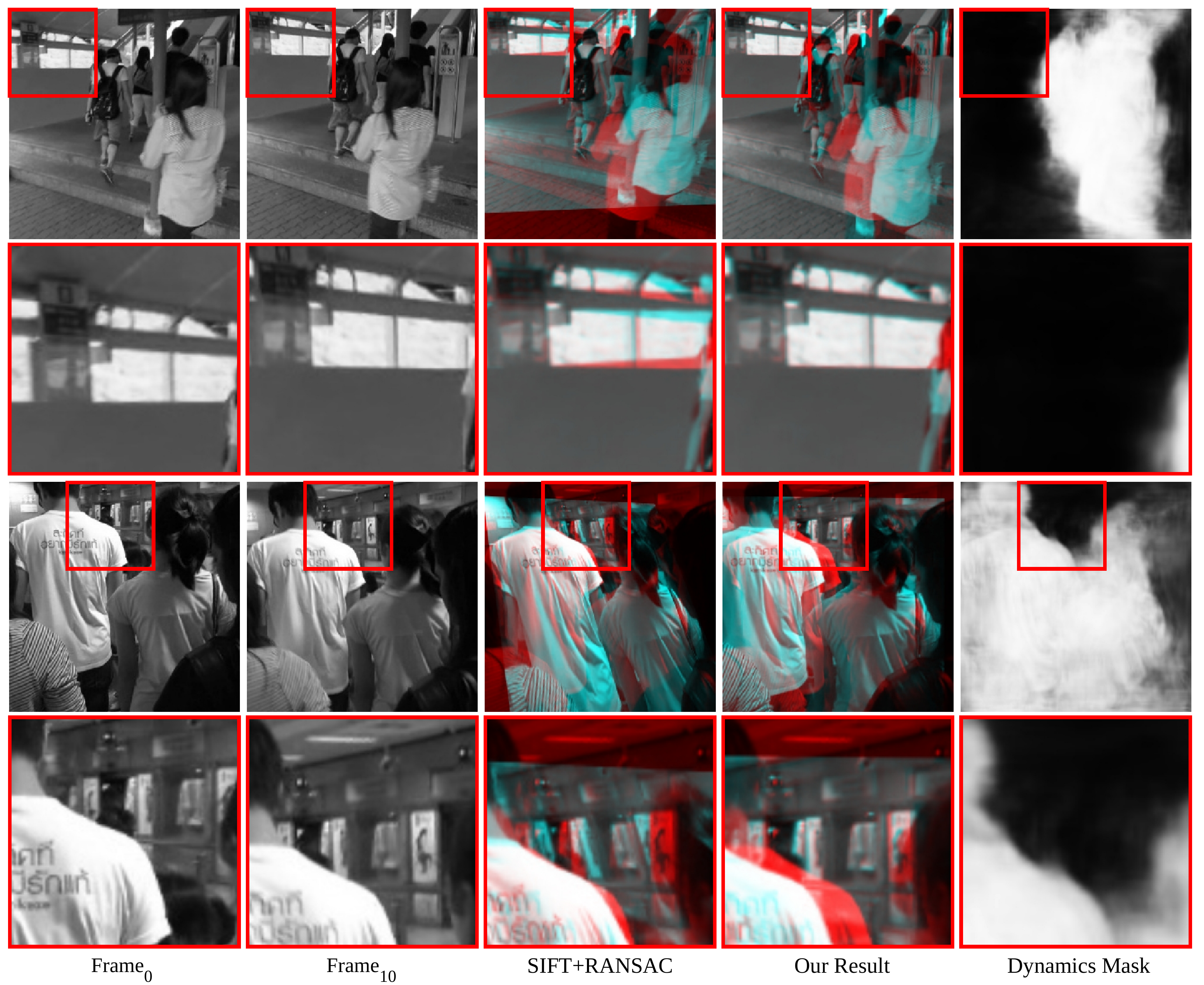}
\end{tabular}
\end{center}
\vspace{-0.3cm}

\caption{Real-world video examples from the NUS video stabilization benchmark~\cite{liu2013bundled}. We estimate homography for pairs of frames that are 10 frames apart. Each homography result is visualized using the red-cyan anaglyph that takes its red channel from Frame$_{10}$ and the cyan channel from the warped Frame$_0$. Any non-boundary colorful pixels \textbf{not on a moving object} indicate misalignment.
\label{fig:real_world}}
\vspace{-0.1in}
\end{figure*}

\noindent\textbf{Real-world videos.} We experimented on videos from the NUS  stabilization benchmark~\cite{liu2013bundled}. We estimate the homography between two frames that are ten frames apart. Fig.~\ref{fig:real_world} shows several challenging examples with significant scene motion and lack of texture in the static background. We visualize each homography estimation result by first using the homography to warp Frame$_0$ to align with Frame$_{10}$ and then creating a red-cyan anaglyph from the warped Frame$_0$ and Frame$_{10}$. Specifically, we take the red channel from Frame$_{10}$ and the cyan channel from the warped Frame$_0$ and merge these channels into an anaglyph image. Any non-boundary colorful pixels not on a moving object indicate misalignment. While our network is trained using synthetic examples (using ground truth homography estimated from images having dynamic objects and static background), it works well with the real videos. Fig.~\ref{fig:real_world} also shows that our network can accurately identify the dynamic content by examining the dynamics masks.

\begin{figure*}
\begin{center}
\begin{tabular}{c}    
    \includegraphics[width=0.82\linewidth]{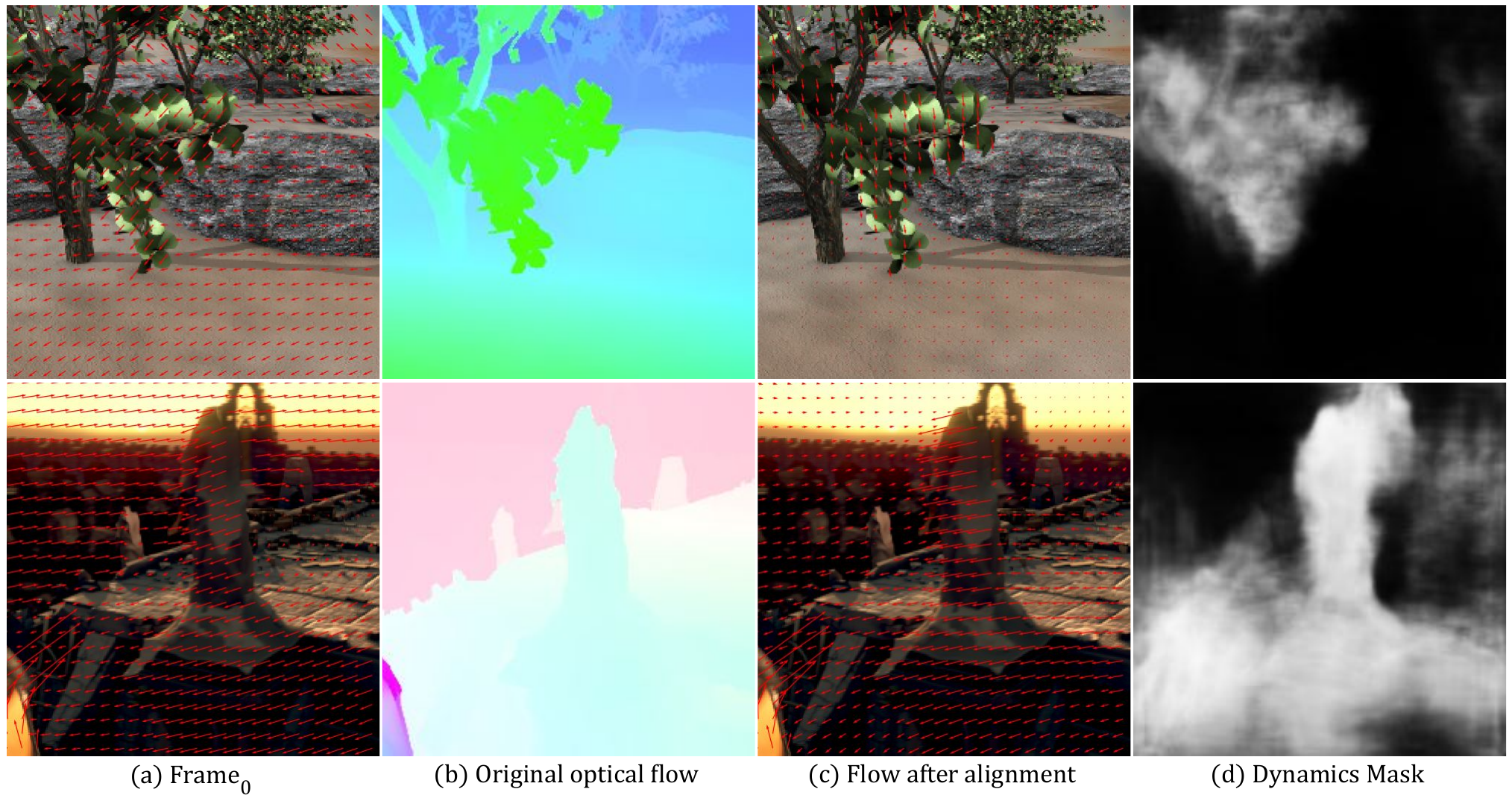}     
\end{tabular}
\end{center}
\vspace{-0.3cm}
\caption{Evaluation on parallax. These examples show that after  a pair of images are aligned, there is little motion in the background as shown in (c). This suggests that our method finds a homography that accounts for the motion in an as-large-as-possible area. As shown in (d), our method treats foreground objects as outliers in the same way as they treat dynamic content. 
\label{fig:exp:parallax_effect_v1}}
\end{figure*}

\noindent\textbf{Parallax.} Although there is no parallax in the static background in our training set, our network can handle parallax well for the above real-world examples. Since no homography in theory can perfectly account for the parallax in an image pair, we examine how our network handles it. To this end, we test our method on examples from optical flow benchmarks, namely Middlebury~\cite{baker2011database} and Sintel~\cite{Butler:ECCV:2012}. In this test, we use our method to estimate the homography between two frames, align them using the estimated homography, and finally compute optical flow between the two aligned frames. As shown in Fig.~\ref{fig:exp:parallax_effect_v1} (c), there is little motion in the background, while the objects that are close to the camera are not aligned. This suggests that our method finds a homography that accounts for the motion in an as-large-as possible area while treating the foreground objects as outliers. As also shown in Fig.~\ref{fig:exp:parallax_effect_v1} (d), our method identifies the foreground object in each example that is far away from the background plane in the dynamics map. Note, while the foreground objects do not move, they are outliers for homography estimation similar to a moving object.

\vspace{-0.05in}
\section{Conclusion}
\vspace{-0.05in}
This paper studied the problem of estimating homography for dynamic scenes. We first collected a large video dataset of dynamic scenes. We developed a multi-scale, multi-stage deep neural network that can handle large global motion and achieve the state-of-the-art homography estimation results when trained and tested on examples derived from the MS-COCO dataset. We further showed that this multi-scale network, when trained on our dynamic-scene dataset, can already handle dynamic scenes to some extent. We then extended this multi-scale network by jointly estimating the dynamics masks and homographies. The dynamics masks enable our method to deal with dynamic scenes better. Our experiments showed that our deep homography neural networks can handle challenging scenarios, such as dynamic scenes, blurriness, and lack of texture.

\paragraph{Acknowledgements.}
Fig.~\ref{fig:figure1} (top), Fig.~\ref{fig:real_world},~\ref{fig:exp:parallax_effect_v1} (top), and Fig.~\ref{fig:exp:parallax_effect_v1} (bottom) originate from MS-COCO~\cite{Lin2014-or}, NUS~\cite{liu2013bundled}, Middleburry~\cite{baker2011database}, and Sintel~\cite{Butler:ECCV:2012} datasets respectively. Fig.~\ref{fig:figure1} (bottom) and Fig.~\ref{fig:dataset:examples} are used under a Creative Commons license from Youtube users Nikki Limo, chad schollmeyer, Lumnah Acres, Liziqi, Dielectric Videos, and 3DMachines. We thank Luke Ding for helping develop our dataset.

{\small
\bibliographystyle{ieee_fullname}
\bibliography{600_bib}
}

\end{document}